\documentclass[review]{elsarticle}
\usepackage{lineno,hyperref}
\usepackage{graphicx}
\usepackage{pdfpages}
\usepackage{times}
\usepackage{epsfig}
\usepackage{amsmath,amssymb,amstext}
\usepackage{amssymb}
\usepackage{algorithm}
\usepackage{algorithmic}

\usepackage{booktabs}
\usepackage{epstopdf}
\usepackage{subfigure}
\usepackage{booktabs}
\usepackage{diagbox}
\usepackage{float}
\usepackage{makecell}
\usepackage{tabularx}
\usepackage{amsthm}

\usepackage{mathrsfs} 
\usepackage{hyperref}
\usepackage{lipsum}
\usepackage{amssymb}
\usepackage{amsmath}
\usepackage{mathrsfs} 
\usepackage{bbm}
\usepackage{booktabs}
\usepackage{threeparttable}
\usepackage{amsfonts}
\usepackage{multirow}
\usepackage{stmaryrd}
\modulolinenumbers[1]
\journal{Neurocomputing}
\begin{document}

\begin{frontmatter}
\title{Multi-Level Graph Contrastive Learning}

\author{Pengpeng Shao$^1$, Tong Liu$^1$, Dawei Zhang$^1$, Jianhua Tao$^{1,2,3}$, Feihu Che$^1$, Guohua Yang$^1$}

\address{$^1$National Laboratory of Pattern Recognition, Institute of Automation, Chinese Academy of Sciences, Beijing, China; \\$^2$School of Artificial Intelligence, University of Chinese Academy of Sciences, Beijing, China;\\ $^3$CAS Center for Excellence in Brain Science and Intelligence Technology, Beijing, China\\}

\begin{abstract}
Graph representation learning has attracted a surge of interest recently, whose target at learning discriminant embedding for each node in the graph. Most of these representation methods focus on supervised learning and heavily depend on label information. However, annotating graphs are expensive to obtain in the real world, especially in specialized domains (i.e. biology), as it needs the annotator to have the domain knowledge to label the graph. To approach this problem, self-supervised learning provides a feasible solution for graph representation learning. In this paper, we propose a Multi-Level Graph Contrastive Learning (MLGCL) framework for learning robust representation of graph data by contrasting space views of graphs. Specifically, we introduce a novel contrastive view - topological and feature space views. The original graph is first-order approximation structure and contains uncertainty or error, while the $k$NN graph generated by encoding features preserves high-order proximity. Thus $k$NN graph generated by encoding features not only provide a complementary view, but is more suitable to GNN encoder to extract discriminant representation. Furthermore, we develop a multi-level contrastive mode to preserve the local similarity and semantic similarity of graph-structured data simultaneously. Extensive experiments indicate MLGCL achieves promising results compared with the existing state-of-the-art graph representation learning methods on seven datasets. 
\end{abstract}


\end{frontmatter}


\section{Introduction}
\label{sec::introduction}

Graphs including millions of nodes and edges are an effective tool to conduct various data analyses for researchers. While implementing the task of data analysis in the original graph is quite intractable due to the inherent high complexity. Currently, the mainstream methods to this obstacle are graph representation learning, which aims to projects the nodes into low-dimensional space and represent them as low dimensional continuous vectors. Traditional graphs representation learning algorithms include random walk-based algorithms (i.e. DeepWalk~\cite{Deepwalk}, Line~\cite{Line}) and Matrix factorization-based algorithms (i.e. DMF~\cite{DMF} and M-NMF~\cite{M-NMF}). With the advent of Graph Neural Networks (GNN), graph representation learning receive more increasingly attention and has been widely applied to various data analytic tasks, such as link prediction, community detection, and anomaly detection on biological networks.

Most graph representation learning methods using GNN focus on supervised fashion and heavily depend on label information to learn the representation of graph-structured data. Unlike common modalities such as image and text, graphs are usually used to represent the relation between knowledge or concepts in specialized domains, graph-structured data need the annotator to have the domain knowledge to label the graph. Thus, the annotating graphs are expensive and limited to obtain in many scientific domains. Therefore, learning the representation of graph-structure data in an unsupervised or self-supervised fashion becomes a more increasingly important task.

Indeed, many reconstruction-based unsupervised algorithms have been proposed for graph representation learning over past years, they either reconstruct the features of graph nodes or the topology structure of the graph, such as DNGR~\cite{DNGR}, SDNE~\cite{SDNE}, NetRA~\cite{NetRA}, and TEA~\cite{TEA}. These methods aim to learn a low-dimensional representation that preserves the locality property and/or global reconstruction capability in autoencoder or variant autoencoder. 
\begin{figure}[http] 
\setlength{\abovecaptionskip}{-1.cm}
\setlength{\belowcaptionskip}{-0.cm} 
\centering
\includegraphics[height=6cm, width=12.7cm]{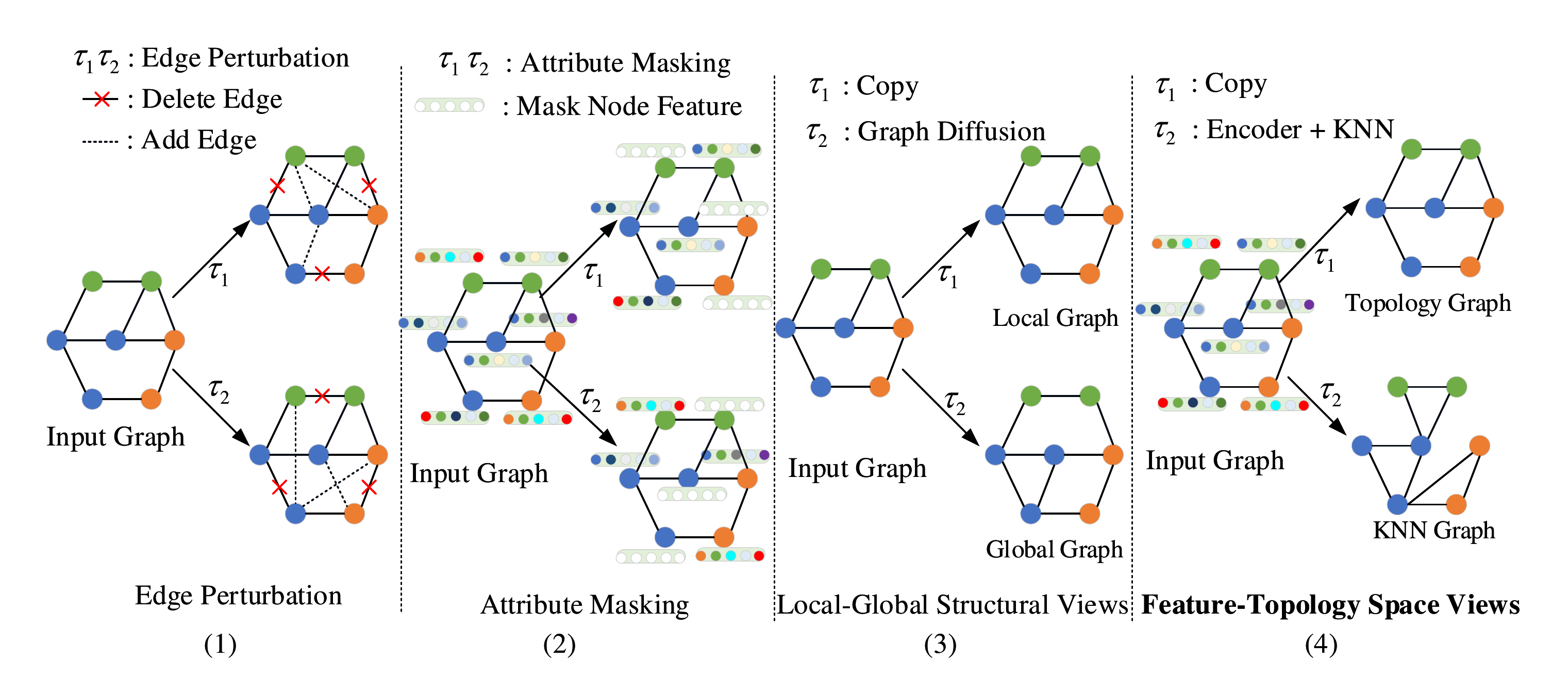}
\caption{Four data augmentations for graphs. The last one is the the proposed data augmentation scheme in this work}
\label{fig::augmentation}
\end{figure}

Inspired by the success of contrastive learning in image understanding (i.e. SimCLR~\cite{SimCLR}, BYOL~\cite{BYOL}, and  MoCo~\cite{MoCo}) and natural language processing (i.e. GPT~\cite{GPT}), contrastive learning coupled with GNNs graph representation learning becomes an open task. The core idea of contrastive learning is designing a pretext task to train an encoder for representation learning, therein, the design of the pretext task includes the design of data augmentation schemes and object function. Thus DGI~\cite{DGI} shuffles node features to augment the training examples, then combines the contrastive objective that maximizing the Mutual information (MI) between the node and global semantic representations to training the feature extractor GNNs. Motivated by multi-view contrastive learning for visual representation learning, MVGRL~\cite{MVGRL} learns node and graph representations by maximizing MI between node representations of one view and graph representation of another view.  Subgraph~\cite{subgraph} learns node representations through MI loss defined based on central nodes and their sampled subgraphs representation.
GCA~\cite{GCA} argues that the data augmentation schemes should preserve intrinsic structures and attributes of graphs and proposes an adaptive augmentation scheme that can extract the important connective structures of the original graph to the augmentation data based on node centrality measures. In addition, GCA~\cite{GCA} also introduces node-level contrastive loss to maximize the agreement between node embeddings in two views. GraphCL~\cite{GraphCL} provides multiple graph augmentations including Edge perturbation, Attribute masking, and maximizes the MI between the node and global semantic representations to learn node representation. Figure~\ref{fig::augmentation} presents four data augmentations for graphs, and Table~\ref{tab:augmentation} gives a comparison of related works in contrastive mode and data augmentation scheme.

\begin{figure}[http] 
\setlength{\abovecaptionskip}{-1.cm}
\setlength{\belowcaptionskip}{-0.cm} 
\centering
\includegraphics[height=6cm, width=13cm]{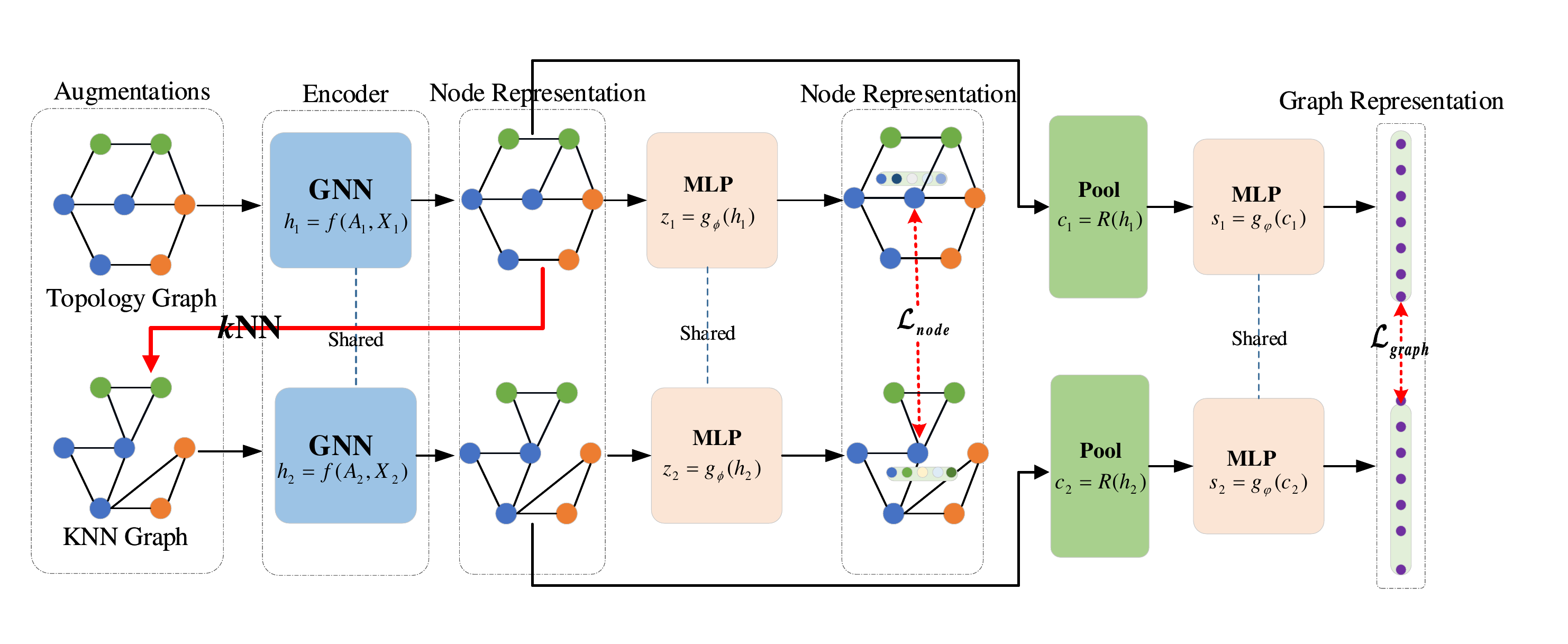}
\caption{The framework of multi-level graph contrastive learning.}
\label{fig::framework}
\end{figure}

Most GNNs models assume that similar nodes are likely to be connected and belong to the same class in the observed graph. However, the observed graph from the real-world contains uncertainty or error, and the nodes are connected in different ways, such as feature similarity, relationship, so that the nodes of different classes may have links. This is inconsistent with the idea of GNN, so that causes unsatisfactory results, a typical example is GNNs perform well on assortative graphs, but do not perform well on disassortative graphs. Therefore, constructing a $k$NN graph by feature similarity or dropping the links between the nodes in different classes are feasible methods. GEN~\cite{GEN} constructs multi-view underlying community structure and attain the
estimated optimal graph in Bayesian inference framework. PTDNet~\cite{PTDNet} prunes task-irrelevant edges by penalizing the number of edges in the sparsified graph with parameterized networks. Inspired by this, we develop a Multi-Level Graph Contrastive Learning (MLGCL) framework for learning robust representation of graph data by contrasting space views of graphs - topological and feature space views. As shown in Figure~\ref{fig::framework}, we apply $k$NN algorithm to encoding features, so that generate the $k$NN graph in the feature space. $k$NN graph not only provides a complementary view but is more suitable to GNN, thus combining both of them in a contrastive learning way can significantly improve the robustness and adaptability of GNNs encoder. Furthermore, we also introduce a multi-level contrastive mode to preserve the local similarity and semantic similarity of graph-structured data simultaneously. Through the proposed novel contrastive views and multi-level contrastive mode, our model MLGCL achieves promising results on seven datasets. 

\begin{table}
\centering
\caption{Comparison with related works in contrast mode and data augmentation scheme}
\label{tab:augmentation}
\begin{tabular}{llllll}
\hline
Methods &  \makecell{Contrastive mode} & \makecell{Views} &  \makecell{Data Augmentation} \\ 
\hline 
DGI & \makecell{Node vs Graph} & \makecell{1} & \makecell{-} \\ 
\hline
GMI & \makecell{Node vs Node} & \makecell{1} & \makecell{-} \\ 
\hline
MVGRL & \makecell{Node vs Graph}& \makecell{2}   & \makecell{Graph Diffusion} \\ 
\hline
SUBG-CON & \makecell{Node vs Graph} & \makecell{1}   & \makecell{-}\\
\hline
GraphCL & \makecell{Node vs Graph} & \makecell{2}  & \makecell{Node Dropping,\\Edge Perturbation, \\Attribute Masking,\\ Extract Subgraph} \\ 
\hline
GCC & \makecell{Graph vs Graph}& \makecell{2}  & \makecell{Extract Subgraph}\\
\hline
GCA&  \makecell{Node vs Node} & \makecell{2}  & \makecell{Adaptive Edge Perturbation, \\Adaptive Attribute Masking } \\ 
\hline
\textbf{Ours}& \makecell{Node vs Node, \\Graph vs Graph} & \makecell{2}  & \makecell{$k$NN Graph Construction}\\ 
\hline
\end{tabular}
\end{table}

\section{Related Work}

\subsection{Graph Representation Learning}
Graph representation learning methods can be broadly classified into three categories: random walk-based algorithms, matrix factorization-based algorithms, and deep learning-based algorithms. Here, we introduce these three types of methods only from the perspective of unsupervised learning.
Random walk-based algorithms, such as DeepWalk~\cite{Deepwalk}, Line~\cite{Line}, Node2vec~\cite{Node2vec}, which is inspired by word2vec algorithm and regards node sequence obtained by random walk as language sentence to analyze. In essence, they also inherit the idea of contrastive learning, that nodes occurring in the same sequence are viewed as the positive sample, and they should have similar representation. Matrix factorization-based algorithms ~\cite{TADW, HOPE}  learns the representation of each node through reconstructing the adjacency matrix or/and fixed-order proximities of the graph. It is worth noting that~\cite{NetMF} proves that random walk-based algorithms can be unified into the matrix factorization framework with closed forms. Early deep learning-based unsupervised algorithms~\cite{DNGR, SDNE} use autoencoder with DNN to learn graph representation. With the advent of GCN, more graph representation learning methods employ powerful GCN as the encoder. For instance, DGI~\cite{DGI}, GraphCL~\cite{GraphCL}, MVGRL~\cite{MVGRL}, and Subgraph~\cite{subgraph}, they all adopt GCN~\cite{GCN}  as encoder and focus on maximizing MI between global graph-level and local node-level embeddings. Different from them, GMI~\cite{GMI} uses a discriminator to directly
measure MI between input and high-level embeddings of a graph neural encoder without data augmentation. GCA~\cite{GCA}  defines a contrastive objective at the node level that distinguishes the embeddings of the same node in these two different views from other node embeddings. Furthermore, GCC~\cite{GCC} proposes a self-supervised GCN pre-training framework, in which it leverages contrastive learning to train GCN~\cite{GCN} on multiple graph datasets to learn the intrinsic and transferable structural representations. GPT-GNN~\cite{GPT-GNN}  also designs a self-supervised attributed graph generation task including attribute generation and edge generation to pre-train GNN.
 
\subsection{Contrastive Learning}
The core idea of the standard contrastive learning approach is designing a pretext task to train a learner model for representation learning. To achieve this pretext task, we first employ prior domain knowledge to automatically generate the training examples that act as pseudo-labels or supervised data of the prediction, then design a contrastive loss function to train the learner model. Therefore, Contrastive Learning consists of three main components: $\emph{Data augmentation schemes}$, generate the supervised data. $\emph{The leaner model}$, which is used to extract data features, commonly consists of an encoder and a head projector. $\emph{Loss function}$, is designed to utilize the augmentation examples to train the learner. Furthermore, how to develop an effective loss function and data augmentation schemes for contrastive learning also is an open problem involving the quality of the leaner. In the following part, we briefly introduce the design of data augmentation and loss function in different domains.

\textbf{Data Augmentation : }
The methods of data augmentation vary in the research domain. In image understanding, the common data augmentation schemes are image rotation, cropping, filtering, add noises, and Mixup, etc. Which can generate two different views of the input that supervise each other. Different from the single-frame image, video is composed of a series of continuous images, which have the characteristics of temporal consistency that adjacent images have similar semantic content. Thus~\cite{Wang} borrows and extends this idea, and employs continuous three images as the contrastive examples,~\cite{Dwi} uses temporal cycle consistency for self-supervision. 

Compared with image and video, graph-structured data have the following priors and properties. First, vertex and edge missing do not alter the graph semantics. Second, the missing partial attributes of each node does not alter the robustness of the graph semantics, Third, the subgraph structure of the graph can hint the full semantics. Fourth, correlated views of the same graph possess semantic consistency, such as local and global views of a graph structure. Thus DGI~\cite{DGI}, GraphCL~\cite{GraphCL}, MVGRL~\cite{MVGRL}, and Subgraph~\cite{subgraph} propose corresponding data augmentations for graphs: node dropping, edge perturbation, attribute masking, extract subgraph, and construct multi-structural views. As illustrated above method that correlated views of the same graph possess semantic consistency, thus MVGRL~\cite{MVGRL} proposes a data augmentation from structure view. In this work, we propose our data augmentation schemes from another view - space view: feature space and topology space views. In these two space, these correlated graphs have similar and intrinsic characteristics, thus the shared learner extracts the similar representation to conduct contrastive learning to learn robust representation for graph data.

\textbf{Loss Function : }
This part summarizes several loss functions that contain contrastive ideas, which are described as follows:

Deepwalk and node2vec extend the idea of skip-gram and propose to optimize the probability loss function that maximizing the probability of nearest vertices of the given a vertex to learn graph representation, the nearest vertices $N_s(u)$ can be obtained by random walk, breadth first search or depth first search algorithm. As illustrated above, these methods also inherit the idea of contrastive learning that nodes occurring in the same sequence are viewed as the positive sample, and they should have similar representation.
\begin{equation}
\label{probabilityloss}
\mathcal{L}_{probability} = \sum_{u \in V}logPr(N_s(u)\vert f(u))
\end{equation}

Self-supervised learning models can be classified as generative, contrastive and generative-contrastive (adversarial). Generative adversarial Network is a typical method of generative-contrastive, the discriminator provides the given true example (supervisory information) to the generator (learner) to boost the capability of the generator. 
\begin{equation}
\label{adversarialloss}
\mathcal{L}_{adversarial} = \underset{G}{\min}~\underset{D}{\max}~\mathbb{E}_{x \sim \mathbb{P}_{data}(x)}[logD(x)] + \mathbb{E}_{z \sim \mathbb{P}_{z}(z)} [log(1 - D(G(z)))]
\end{equation}
Where $z$ is noise, $D(\cdot)$ and $G(\cdot)$ correspond to discriminator and generator, respectively, 

Triplet loss plays a very important role in deep metric learning : Given an anchor $x$, a positive $x_p$ of the same class as the anchor, a negative $x_f$ of a different class, the triplet loss targets at achieving this case that the distance of $x$ and $x_f$ greater than the distance of $x$ and $x_p$ plus margin $m$.  
\begin{equation}
\label{tripletloss}
\mathcal{L}_{triplet} = max(d(x,x_p) - d(x, x_f) +m)
\end{equation}

Currently, standard contrastive loss can be divided into two categories, one is maximizing the similarity of the corresponding nodes between two views, the other is maximizing the MI between node and graph representation.
\begin{equation}
\label{ContrastiveLoss}
\mathcal{L}_{node}(z_i, z_j) = log\frac{exp((z_i^a)^{T}z_i^b/\tau)}{\sum_{j=1,j\neq i}^{K}exp((z_i^a)^{T}z_i^b/\tau) + exp((z_i^a)^{T}z_j^a/\tau) + exp((z_i^a)^{T}z_j^b/\tau)}
\end{equation}
\begin{equation}
\label{MutualLoss}
\mathcal{L}_{mutual} = \frac{1}{2K} \sum_{i=1}^{K}[MI(z_i^a, s^b) + MI(z_i^b, s^a)]
\end{equation}
Where, $z$ denotes node representation, $s$ is graph representation, $a$ and $b$ denotes two views, respectively. the $\mathcal{L}_{contrastive}$ contains a pair of positive examples $(z_i^a, z_i^b)$ and $K$-1 pair of negative examples $(z_i^a, z_j^a)$ and $(z_i^a, z_j^b)$.  We randomly sample a minibatch of $K$ examples, we can derive $2K$ data points by conducting one data augmentation. Given a positive pair, we treat the other $2(K-1)$ examples as negative examples. $\tau$ is a temperature hyperparameter that is used to smooth the results.

\section{Prelininaries} \label{Prelininaries}
Prior to going further, we first introduce the preliminary concepts used in the following sections. Let $G(A, X)$ denotes a undirected graph, where $V = \{v_1, v_2, \cdots, v_N\}$ is node set, $N = \vert V \vert$ is the number of graph node. $E$ denotes edge set. The adjacent matrix $A \in R^{N \times N}$  represents the topological structure of graph $G$, its elements suggest the strength of connection between $v_i$ and $v_j$. $X \in R^{N \times F}$ suggests the matrix consisting of the content features associated with each node. Contrastive learning aims to learn an encoder model $f(\cdot)$ which can learn a robust representation for each node $v_i$ that is insensitive to the perturbation caused by data augmentations $\tau$. Here, we denote the node representation learned by the encoder model as $Z = f(A, X)$. The learned representation then be applied to the downstream tasks, such as node classification.

\section{Multi-Level Graph Contrastive Learning Framework}

To learn a robust graph representation in a self-supervised fashion, we propose a Multi-Level Graph Contrastive Learning (MLGCL) framework as Figure~\ref{fig::framework}. To be specific, we first sample a pair of graph augmentation functions $\tau_1$ and $\tau_2$ from augmentation pool $\tau$, which is applied to the input graph to generate the augmented graph of two views. Then we use a pair of shared GNN-based encoder to extract node representation, and further employ the pooling layer to extract the graph representation. After that, we utilize a shared MLP layer to project the nodes representation from both views into the space where the node-level contrastive loss is computed. Similarly, we also project the graphs representation from both views into the space where the graph-level contrastive loss is computed. Finally, we learn the parameters of the encoder through optimize the proposed multi-level loss function. Note that, the augmentation graph of the second view is generated by the encoding feature, thus $\tau_2$ includes an encoder and $k$NN function.

Thus, MLGCL is mainly composed of the following components:  
\begin{itemize}
\item Graph Data Augmentation : Its goal is to exert perturbation on the input graph to generate two correlated graphs of the same graph. In this work, we extract augmentation graph structure from space view to conduct contrastive learning. 
\item GNN Encoder: GNN-based Encoder $f(\cdot)$ is used to learn node representation $z_1$, $z_2$ for two augmented graphs. In the inference phrase, we only use GNN encoder to learn node representation for downstream tasks.
\item MLP : The head projection MLP layer $g(\cdot)$ maps the representation to the space in which the contrastive loss is computed.
\item Graph Pooling : The graph pooling layer (i.e. readout) $R(\cdot)$ is employed to learn graph representation.
\item Loss Function : The multi-level loss function is proposed to preserve the low-level ``local'' and high-level ``global'' agreement, simultaneously. 
\end{itemize}

\subsection{Data Augmentation for Graphs}
In visual representation learning, contrasting two views of the image allows encoders to learn their common and discriminant representations. Which inspires us to seeks to two different and correlated views of the graph. In this work, we extract two correlated augmentation graph structure from space view to conduct contrastive learning. Given the graph structure of topology space $G(A, X)$, we first employ GNN encoder to extract the encoding features $Z$ of topology graph, then apply $k$-nearset neighbour to $Z$ to construct $k$NN graph with community structure $G_f(A_f, X)$, where $A_f$ is the adjacency matrix of $k$NN graph. Therein, constructing the feature map using $k$NN can described as two steps: First, calculating the similarity matrix $S$ based on $N$ encoding features $Z$. Second, choosing top $k$ similar node pairs for each node to set edges and finally get the adjacency matrix of $k$NN graph $A_f$. In fact, there are many schemes to obtain similarity matrix $S$, such as distance-based Similarity Calculation Method (i.e. Euclidean distance, Mahalanobis distance),  Cosine-based similarity calculation (i.e. Cosine similarity, Pearson correlation), and Kernel-based similarity calculation (i.e. Gaussian kernel, Laplace kernel). We list three popular ones here, in which $x_i$ and $x_j$ are feature vectors of nodes $i$ and $j$:

1) \textbf{Mahalanobis Distance}:
\begin{equation}
\label{Mahalanobis distance}
{ \emph S_{ij}} = \sqrt{(x_i -x_j)^{T}M(x_i -x_j)}
\end{equation}
Where $M$ is a positive semidefinite matrix, which plays the role of the inverse covariance matrix

2) \textbf{Cosine Similarity} : It uses the cosine value of the angle between two vectors to measure the similarity.
\begin{equation}
\label{CosineSimilarity}
{ \emph S_{ij}} = \frac{x_i \cdot x_j}{\vert x_i \vert \vert x_j\vert}
\end{equation}

3) \textbf{Gaussian Kernel} : The similarity is calculated by the Eq.~\ref{HeatKernel} where $\sigma$ is the kernel width of Gaussian kernel. 
\begin{equation}
\begin{aligned}
\label{HeatKernel}
S_{ij} = e^{-\frac{\| x_i - x_j\|^2}{2 \sigma^2}}
\end{aligned} 
\end{equation}
Here, we uniformly choose the Cosine Similarity to obtain the similarity matrix $S$.

\subsection{Encoder}
Given topology graph $G(A, X)$ and $k$NN graph $G_f(A_f, X)$, we employ a two-layer GCN as our encoder model to obtain their latent node representations matrix, repectively. The encoder $f(\cdot)$ can be formally represented as 
\begin{equation}
\begin{aligned}
\label{GCN}
Z^{l+1} = f(A, X) = \sigma(\widetilde{A} Z^l W^{l})
\end{aligned} 
\end{equation}
Here, $\widetilde{A} = \widehat{D}^{-1/2}\widehat{A}\widehat{D}^{-1/2}$ is the symmetrically normalized adjacency matrix, $\widehat{A} = A + I_{n}$ is the adjacency matrix with self-loops, $I_{n} \in \mathbb{R}^{n \times n}$ is the identity matrix, and $\widehat{D} = \Sigma_{j}\widehat{A}_{ij}$ is degree matrix. $Z^l$ contains the embedding (row-wise) of the graph vertices in the $l$-th layer, and $Z^{0}= X$ is the input for GCN in the first layer.  $W^{l}$ is the parameters matrix, and $\sigma(\cdot)$ is nonlinearity activation function (i.e., ReLU, Sigmoid).

For node representation $Z_a$, $Z_b$ of each view, we utilize a graph pooling layer $P(\cdot)$ : $\mathbb{R}^{N \times d} \longmapsto \mathbb{R}^{d}$ (i.e. readout function) to derive their graph representation:
\begin{equation}
\begin{aligned}
\label{graphrepresentation}
c = P(H) = \sigma(\frac{1}{N}\sum_{i=1}^{N}h_i)
\end{aligned} 
\end{equation}
In addition, in order to conduct corresponding node representation and graph representation contrast of two views, we employ MLP layer $g_{\phi}(\cdot)$ and $g_{\varphi}(\cdot)$ : $\mathbb{R}^{N \times d} \longmapsto \mathbb{R}^{N \times d}$ to project node and graph representation into the space where the contrastive loss is computed, respectively.

\subsection{Multi-Level Loss Function}
The core idea of contrastive learning is designing an appropriate pretext task to train an encoder model to extract rich node and graph representation.
Here, we employ multi-level loss function to update the parameters of the encoder. Multi-level loss function, namely, the joint contrastive loss function from local to global, consists of two parts: the contrast of low-level node representation between two views, and the contrast of high-level graph representation between two views.

Given positive pair $(z_i, z_j)$, the node-level contrastive loss function is defined as,
\begin{equation}
\label{ContrastiveLoss1}
\mathcal{L}_{node}(z_i^a, z_i^b) = log\frac{exp((z_i^a)^{T}z_i^b/\tau)}{\sum_{j=1,j\neq i}^{K}exp((z_i^a)^{T}z_i^b/\tau) + exp((z_i^a)^{T}z_j^a/\tau) + exp((z_i^a)^{T}z_j^b/\tau)}
\end{equation}
Due to the two views is symmetric, the loss of the other view is defined as $\mathcal{L}_{node}(z_i^b, z_i^a)$. Thus we maximize the agreement of the nodes between two views through optimizing the followings,  
\begin{equation}
\label{ContrastiveLoss2}
\mathcal{L}_{node} = \mathcal{L}_{node}(z_i^a, z_i^b) +  \mathcal{L}_{node}(z_i^b, z_i^a)
\end{equation}


Similarly, given positive examples $(s^a, s^b)$ and negative examples $(s^a, \tilde{s}^a)$,  $(s^a, \tilde{s}^b)$,  the graph representation contrast between two views is defined as, 
\begin{equation}
\label{graphLoss1}
\mathcal{L}_{graph}(s^a, s^b) = log\frac{exp((s^a)^{T}s^b/\tau)}{exp((s^a)^{T}s^b/\tau) + exp((s^a)^{T}\tilde{s}^a/\tau) + exp((s^a)^{T}\tilde{s}^b/\tau)}
\end{equation}
Here, to generate negative graph samples, we randomly shuffle the features to derive negative adjacent matrix $\tilde{A}$ and $\tilde{A}_f$, and then obtain negative samples $(s^a, \tilde{s}^a)$,  $(s^a, \tilde{s}^b)$. The loss for another view is defined as $\mathcal{L}_{graph}(s^b, s^a)$. Thus, the overall graph representation contrast, formally given by, 
\begin{equation}
\label{graphLoss2}
\mathcal{L}_{graph} = \mathcal{L}_{graph}(s^a, s^b) + \mathcal{L}_{graph}(s^b, s^a)
\end{equation}

Finally, through combining node-level contrastive loss with graph-level contrastive loss, the multi-level loss of our model to be maximized
can be presented as,
\begin{equation}
\label{Loss}
\mathcal{L} = \mathcal{L}_{node} + \lambda \mathcal{L}_{graph}
\end{equation}
where $ \lambda$ is tuning parameters to balance the importance of $\mathcal{L}_{node}$ and $\mathcal{L}_{graph}$, respectively.

\section{Experiment Result}

\subsection{Datasets } \label{dataetEvalMetrics}
To evaluate the effectiveness of the representation learned by the proposed model, we conduct extensive experiments on multiple datasets from different domains. Here, We use seven datasets Cora, Citeseer, Pubmed, WikiCS, CoauthorCS, Cornell, and Actor to demonstrate the scalability of our methods. Therein, Cora, Citeseer, and Pubmed are three citation networks in which the vertexes are academic papers and the edges represent the citation relationship between papers. WikiCS is a graph dataset of articles about computer science, where the nodes refer to articles, edges are hyperlinks between articles. CoauthorCS is a coauthor network in which the two nodes that have a connection relationship represent two articles with the same author. Cornell is a web pages network where nodes are web pages, edges are links between web pages. Actor is a cooccurence network in which the nodes refer to actors, each edge represents the two actors that occur on the same Wikipedia page. The detailed information of all these datasets is provided in Table~\ref{tab:dataset}.
\begin{table*}
\centering
\label{tab:dataset}
\begin{tabular}{lllllll}
\hline
Dataset & \makecell{Nodes} & \makecell{edges}  & \makecell{Features} & \makecell{Labels} & \makecell{Domain}\\ 
\hline 
Cora  & \makecell{2,708} & \makecell{5,429}  & \makecell{1,433} & \makecell{7} & \makecell{Citation Networks}\\ 
Citeseer & \makecell{3,312} & \makecell{4,732}  & \makecell{3,703} & \makecell{6} & \makecell{Citation Networks}\\ 
PubMed & \makecell{19,717} & \makecell{44,338} & \makecell{500} & \makecell{3} & \makecell{Citation Networks}\\ 
WikiCS & \makecell{11,701} & \makecell{216,123}  & \makecell{300} & \makecell{10} & \makecell{Hyperlinks Networks}\\ 
CoauthorCS & \makecell{18,333} & \makecell{81,894}  & \makecell{6,805} & \makecell{15} & \makecell{Coauthor Networks}\\
Cornell  & \makecell{183} & \makecell{280}  & \makecell{1,703} & \makecell{5} & \makecell{Web Pages Networks }\\
Actor & \makecell{7,600} & \makecell{26,752}  & \makecell{932} & \makecell{5} & \makecell{Cooccurence Networks}\\
\hline
\end{tabular}
\caption{Graph Datasets}
\end{table*}

\subsection{Evaluation Protocol} \label{Evaluation Protocol}

In this work, we conduct node classification experiments and employ the experimental protocol of the previous state-of-the-art approaches to evaluate the performance of the proposed method. Specifically, we use the node representation learned by our model to train the logistic regression classifier, and then give the classification results on the test nodes. Followed DGI, we adopt the mean classification accuracy with standard deviation to assess the performance for all these datasets.

\subsection{Baselines} \label{Baselines}
To assess the proposed method, we employ the following two categories methods as our baselines: (1) unsupervised learning methods,  including the naive method that use a logistic regression classifier on raw input node features, classical methods DeepWalk~\cite{Deepwalk} and the DeepWalk-F that concatenates the learned embeddings with input node features, unsupervised version of GraphSage~\cite{GraphSage} (abbreviated as Unsup-GraphSAGE), self-supervised methods DGI~\cite{DGI}, GMI~\cite{GMI}, Sub-graph~\cite{subgraph}, MVGRL~\cite{MVGRL}, GraphCL~\cite{GraphCL}, and GCA~\cite{GCA}. (2) supervised learning methods MLP~\cite{GATorMLP}, Label Propagation (LP)~\cite{LP}, PLANETOID~\cite{PLANETOID}, CHEBYSHEV~\cite{CHEBYSHEV}, GCN~\cite{GCN}, GAT~\cite{GATorMLP}, and supervised version of GraphSage. Notably, for all baselines, we report their performance based on their official implementations or choose optimal hyper-parameters carefully after reproducing
the code for different baselines in this paper to ensure the fairness of comparison experiments.

\subsection{Implementation Details} \label{impleDetails}

The proposed model is implemented on Pytorch~\cite{pytorch} framework and geometric deep learning extension library, and it is optimized by Adam algorithm~\cite{Adam}. The parameters in the proposed model are initialized by Xavier initialization. To have fair comparisons, we follow MVGRL and 
set the number of GCN layers and the number of epochs and to 1 and 2,000, respectively. Here, we set the size of the latent dimension of both node and graph representations to 512. To choose the best classifier model for the test, we also utilize early stopping with a patience of 20.
\subsection{Comparison with Baseline Algorithms} \label{compSOTA}

Table~\ref{tab:comparsionresults1} and Table~\ref{tab:comparsionresults2} present the classification results of the proposed model and baseline models on seven graph datasets, and we have the following observations and analyses. First, on the whole, the proposed model achieves considerable performance compared with baseline models. For instance, our method outperforms the state-of-the-art self-supervised methods GCA and GMI 4$\%$ and 1.4$\%$ on Cora dataset, respectively. On Cornell dataset, our method outperforms MVGRL and GCA 4.7$\%$ and 13.4$\%$, respectively. Second, although without the guidance of the label information, our method obtains superior performance on the seven datasets compared with supervised learning methods including GCN, GAT, and GraphSage. This verifies that the pseudo-label information generated by data augmentation exerts a powerful effect on self-supervised representation learning. Third, GRACE proposes node-level contrastive mode, while GCA which is an improved version of GRACE designs an adaptive data augmentation based on GRACE to learn representations that are insensitive to perturbation on unimportant nodes and edges. However, we can obtain that GCA does not obtain explicit improvement than GRACE from the results presented in Table~\ref{tab:comparsionresults1} and Table~\ref{tab:comparsionresults2}. We argue that the main reasons are as follows. (1)  That the importance of an edge is determined by its centrality degree is not universally accurate, such as some nodes have only one edge. (2) Even though removing the important edge of the node, the performance of the model may not much worse than that of GCA, because removing edges uniformly on the whole graph does not affect the semantics of the node. Fourth, there are the following two differences between GCA and our method: (1) GCA designs a topology level augmentation scheme based on edge centrality measures and an attribute level augmentation scheme by adding more noise on node features, respectively. While our method employs the topological graph of topological space and the community graph of feature space as contrastive examples of two views. (2) Our method develops the node-level contrastive mode and proposes a multi-level contrastive mode from local and global views. These two designs also enable our method to outperform GCA on all datasets. Fifth, MVGRL is the closest work to our method,  MVGRL designs the augmentation examples from local and global views, while the proposed method develops contrastive mode from local and global views. From the overall experimental results, our method has a similar classification performance to MVGRL. Finally, 
the considerable performance of our model verifies that our proposed data augmentation scheme and multi-level contrastive mode is capable of helping boost graph representation quality.

\begin{table}
\centering
\caption{Summary of performance on node classification in terms of accuracy in percentage with standard deviation. Available
data for each method during the training phase is presented in the second column, where A, X, Y correspond to the
adjacency matrix, node features, and node labels, respectively. OOM indicates running Out-Of-Memory on a 16GB GPU. For clarity, the champion performance is highlighted in boldface, the runner-up performance is underlined.}
\label{tab:comparsionresults1}
\begin{tabular}{llllll}
\hline
Algorithm &  Input data & \makecell{Cora} &  \makecell{Citeseer} &  \makecell{Pubmed} \\ 
\hline 
Raw features & \makecell{X} & 47.9 $\pm$ 0.4 & 49.3 $\pm$ 0.2 & 69.1 $\pm$ 0.2 \\   
DeepWalk & \makecell{A} & \makecell{67.2} & \makecell{43.2} & \makecell{63.0} \\  
DeepWalk-F & \makecell{A, X} & 70.7 $\pm$ 0.6 & 51.4 $\pm$ 0.5 & 74.3 $\pm$ 0.9 \\  
Unsup-GraphSAGE & \makecell{A, X}  & 75.2 $\pm$ 1.5 & 59.4 $\pm$ 0.9 & 70.1 $\pm$ 1.4 \\  
DGI & \makecell{A, X} & 82.3 $\pm$ 0.6 & 71.8 $\pm$ 0.7 & 76.8 $\pm$ 0.6 \\  
GMI & \makecell{A, X} & 83.0 $\pm$ 0.3 & 73.0 $\pm$ 0.3 & 79.9 $\pm$ 0.2 \\  
SUBG-CON & \makecell{A, X} & 83.5 $\pm$ 0.5 & 73.2 $\pm$ 0.2 & \underline{81.0 $\pm$ 0.1} \\  
MVGCL & \makecell{A, X} & \textbf{85.6 $\pm$ 0.5} & \underline{73.3 $\pm$ 0.5} & 80.1 $\pm$ 0.7  \\  
GraphCL-Edge & \makecell{A, X} & 81.4 $\pm$ 0.4 & 72.5 $\pm$ 0.5 & 77.4 $\pm$ 0.4 \\
GraphCL-Mask & \makecell{A, X} & 81.5 $\pm$ 0.5 & 72.7 $\pm$ 0.5 & 77.8 $\pm$ 0.6\\ 
GRACE & \makecell{A, X} & 80.0 $\pm$ 0.4 & 71.7 $\pm$ 0.6 & 79.5 $\pm$ 1.1 \\ 
GCA & \makecell{A, X} & 80.5 $\pm$ 0.5 & 71.3 $\pm$ 0.4 & 78.6 $\pm$ 0.6   \\   
\hline
\textbf{MLGCL} & \makecell{A, X} & \underline{84.4 $\pm$ 0.4} & \textbf{73.5 $\pm$ 0.5} & \textbf{81.3 $\pm$ 0.5} \\  
\hline
MLP & \makecell{X, Y} & \makecell{55.1} & \makecell{46.5} & \makecell{71.4} \\
LP & \makecell{A, Y} & \makecell{68.0} & \makecell{45.3} & \makecell{63.0} \\
PLANETOID & \makecell{X, Y} & \makecell{75.7} & \makecell{64.7} & \makecell{77.2} \\
CHEBYSHEV & \makecell{A, X, Y} & \makecell{81.2} & \makecell{69.8} & \makecell{74.4} \\
GCN &\makecell{A, X, Y} & \makecell{81.5} & \makecell{70.3} & \makecell{79.0} \\
GAT & \makecell{A, X, Y} & 83.0 $\pm$ 0.7  & 72.5 $\pm$ 0.7  & 79.0 $\pm$ 0.3 \\
GraphSAGE &\makecell{A, X, Y} & 79.2 $\pm$ 1.5 & 71.2 $\pm$ 0.5 & 73.1 $\pm$ 1.4 \\
\hline
\end{tabular}
\end{table}

\begin{table}
\centering
\caption{Summary of performance on node classification in terms of accuracy in percentage with standard deviation. Available
data for each method during the training phase is presented in the second column, where A, X and Y correspond to the
adjacency matrix, node features and node labels, respectively. For clarity, the champion performance is highlighted in boldface, the runner-up performance is underlined.}
\label{tab:comparsionresults2}
\begin{tabular}{llllll}
\hline
\!\!\!\!Algorithm \!\!\!\!\!&\!\!\!\!\!  Input data \!\!\!\!\!&\!\!\!\!\! \makecell{WikiCS} \!\!\!\!\!&\!\!\!\!\!  \makecell{CoauthorCS} \!\!\!\!\!&\!\!\!\!\! \makecell{Cornell} \!\!\!\!\!&\!\!\!\!\!  \makecell{Actor}\\ 
\hline 
\!\!\!\!Raw features & \makecell{X} & 71.98 $\pm$ 0.00 & 90.37 $\pm$ 0.07 & 57.60 $\pm$ 3.2 & 25.62 $\pm$ 0.5 \\ 
\!\!\!\!DeepWalk & \makecell{A} & 74.35 $\pm$ 0.06 & 84.61 $\pm$ 0.22 & 56.72 $\pm$ 4.6 & 27.81 $\pm$ 0.8\\
\!\!\!\!DeepWalk-F  & \makecell{A, X} & 77.21 $\pm$ 0.03 & 87.70 $\pm$ 0.04 & 58.26 $\pm$ 0.2 & 48.56 $\pm$ 0.3\\  
\!\!\!\!MVGRL & \makecell{A, X} & 77.52 $\pm$ 0.08 & 92.11 $\pm$ 0.12 & 61.19 $\pm$ 3.0& \underline{30.52 $\pm$ 0.7} \\
\!\!\!\!DGI & \makecell{A, X} & 75.35 $\pm$ 0.14 & 92.15 $\pm$ 0.63 & 55.68 $\pm$ 4.8& 26.80 $\pm$ 0.8 \\
\!\!\!\!GMI & \makecell{A, X} & 74.85$\pm$ 0.08 & \makecell{OOM} & \underline{61.35 $\pm$ 6.0} & 29.51 $\pm$ 0.8\\
\!\!\!\!\!GRACE & \makecell{A, X} & 78.19 $\pm$ 0.01 & 92.93 $\pm$ 0.01 & 53.78 $\pm$ 7.2 & 27.06 $\pm$ 1.1\\   
\!\!\!\!GCA & \makecell{A, X} & \underline{78.35 $\pm$ 0.01} & \textbf{93.10 $\pm$ 0.01} & 52.47 $\pm$ 5.4  & 26.38 $\pm$ 0.4 \\
\hline
\!\!\!\!\textbf{MLGCL} & \makecell{A} & \textbf{78.63 $\pm$ 0.05} & 92.86 $\pm$ 0.01 & \textbf{61.89 $\pm$ 6.2} & \textbf{30.74 $\pm$ 0.7}\\
\hline
\!\!\!\!Chebyshev & \makecell{A, X, Y} & 75.63 $\pm$ 0.60  & 91.52 $\pm$ 0.00 & 55.81 $\pm$ 4.6 & 29.64 $\pm$ 0.5\\
\!\!\!\!GCN & \makecell{A, X, Y} & 77.19 $\pm$ 0.12& \underline{93.03 $\pm$ 0.35} & 56.78 $\pm$ 4.5 & 30.28 $\pm$ 0.6\\
\!\!\!\!GAT & \makecell{A, X, Y} & 77.65 $\pm$ 0.11 &92.31 $\pm$ 0.24 & 59.21 $\pm$ 2.3 & 26.30 $\pm$ 1.3\\
\hline
\end{tabular}
\end{table}

\subsection{Ablation Study}

In this section, we conduct two ablation studies to study the impact of each component of the proposed model. To study the impact of our data augmentation, We investigated another three data augmentation schemes including Edge Perturbation, Attribute Masking, and Graph Diffusion. Similarly, we 
drop the component of multi-level contrastive mode to study its impact. The results are presented in Table~\ref{tab:AblationStudy}.
\renewcommand{\arraystretch}{1.2} 
\begin{table}[tp]
\centering
\caption{Performance comparison of variants of the proposed model}
\label{tab:AblationStudy}
\begin{tabular}{llllll}
\hline
Algorithm &  Available data & \makecell{Cora} &  \makecell{Citeseer} &  \makecell{Pubmed} \\ 
\hline 
MLGCL-edge & \makecell{A, X} & 68.9 $\pm$ 0.3& 60.1 $\pm$ 0.3 & 64.3 $\pm$ 0.4 \\  
MLGCL-mask & \makecell{A, X} & 70.9 $\pm$ 0.4& 61.3 $\pm$ 0.4 & 65.4 $\pm$ 0.4 \\ 
MLGCL-L\&G & \makecell{A, X} &80.8 $\pm$ 0.5& 70.1 $\pm$ 0.5 & 75.6 $\pm$ 0.7\\      
\hline
\textbf{MLGCL} & \makecell{A, X} & \textbf{84.4 $\pm$ 0.4}& \textbf{73.5 $\pm$ 0.5} & \textbf{80.5 $\pm$ 0.5} \\  
\hline
MLGCL-node vs node & \makecell{A, X} & \makecell{80.9} & \makecell{70.7} &\makecell{77.5} \\  
MLGCL-graph vs graph & \makecell{A, X} & \makecell{80.4}  & \makecell{71.8} &\makecell{78.2} \\  
\hline
\end{tabular}
\end{table}

\subsubsection{Impact of Data Augmentation}
As illustrated above, we investigated another three data augmentation schemes, Edge Perturbation, Attribute Masking, and Graph Diffusion. Thus in Table~\ref{tab:AblationStudy}, MLGCL-edge denotes the model that replacing the proposed data augmentation scheme with Edge Perturbation augmentation. Similarly, MLGCL-mask and MLGCL-L\&G  refer to the model that replaces the proposed data augmentation scheme with Attribute Masking and Graph Diffusion augmentations, respectively. The results of the upper part of Table~\ref{tab:AblationStudy} show that the proposed augmentation scheme improves model performance on Cora, Citeseer, and Pubmed datasets, and our proposed model gains 15.3$\%$ absolute improvement compared to MLGCL-edge. The results verify the effectiveness of our data augmentation schemes on feature space.

\subsubsection{Impact of Contrastive Mode}
We considered three contrastive modes including, node va node, graph vs graph, and these two modes are the components of the proposed contrastive mode. Thus MLGCL-node vs node suggests the model that replacs the proposed multi-level contrastive mode with node vs node contrastive mode, MLGCL-graph vs graph denotes the model that replacing the proposed multi-level contrastive mode with graph vs graph contrastive mode. The results of the lower part in Table~\ref{tab:AblationStudy} suggest that the proposed multi-level contrastive mode is superior to its component mode, further indicate that local and global contrastive mode can collaborate to achieve better performance. Please kindly note that the above results in Table~\ref{tab:AblationStudy} only verify that the proposed method is effective and achieve the best performance compared to the variants of the proposed method under our settings, it can not suggests the proposed data augmentation scheme or contrastive mode is best universally.


\section{Conclusion}\label{Conclusion}
 This work introduces a novel contrastive learning framework for graph representation learning. First, due to original graph is first-order approximation structure and contains uncertainty or error, while $k$NN graph with community structure that generated by encoding features preserves high-order proximity, thus $k$NN graph not only provide a complementary graph structure from space view but also is more suitable to GNN encoder. Furthermore, we develop a multi-level contrastive mode to preserve the local similarity and semantic similarity of graph-structured data simultaneously. Extensive experiments indicate the proposed model achieves considerable results compared with the existing graph representation learning methods on seven datasets.

\section*{References}
\bibliography{mybibfilespp}
\end{document}